\documentclass[letterpaper, 10 pt, conference]{ieeeconf}  
\usepackage{hyperref}

\IEEEoverridecommandlockouts                              

\usepackage{graphics} 
\usepackage{epsfig} 
\usepackage{mathptmx} 
\usepackage{multirow}
\usepackage{times} 
\usepackage{amsmath} 
\usepackage{amssymb}  

\title{\LARGE \bf
CapsDT: Diffusion-Transformer for Capsule Robot Manipulation
}

\author{Xiting He, Mingwu Su, Xinqi Jiang, Long Bai, Jiewen Lai, and Hongliang Ren, \textit{Senior Member, IEEE}
\thanks{*This work was supported by Hong Kong RGC CRF C4026-21GF, GRF (14203323, 14216022, \& 14211420), NSFC/RGC Joint Research Scheme N\_CUHK420/22, Shenzhen-HK-Macau Technology Research Programme (Type C) STIC Grant 202108233000303. (Corresponding to: H. Ren)}
\thanks{X. He, M. Su, X. Jiang, L. Bai, J. Lai, and H. Ren are with the Department of Electronic Engineering, The Chinese University of Hong Kong (CUHK), Hong Kong, China; and also with the CUHK Shenzhen Research Institute, Shenzhen, China. (E-mail: xitinghe@cuhk.edu.hk, hlren@ee.cuhk.edu.hk)}
}

\begin{document}

\maketitle
\thispagestyle{empty}
\pagestyle{empty}

\begin{abstract}

Vision-Language-Action (VLA) models have emerged as a prominent research area, showcasing significant potential across a variety of applications. However, their performance in endoscopy robotics, particularly endoscopy capsule robots that perform actions within the digestive system, remains unexplored. The integration of VLA models into endoscopy robots allows more intuitive and efficient interactions between human operators and medical devices, improving both diagnostic accuracy and treatment outcomes. In this work, we design CapsDT, a Diffusion Transformer model for capsule robot manipulation in the stomach. By processing interleaved visual inputs, and textual instructions, CapsDT can infer corresponding robotic control signals to facilitate endoscopy tasks. In addition, we developed a capsule endoscopy robot system, a capsule robot controlled by a robotic arm-held magnet, addressing different levels of four endoscopy tasks and creating corresponding capsule robot datasets within the stomach simulator. Comprehensive evaluations on various robotic tasks indicate that CapsDT can serve as a robust vision-language generalist, achieving state-of-the-art performance in various levels of endoscopy tasks while achieving a 26.25\% success rate in real-world simulation manipulation.
\end{abstract}

\section{INTRODUCTION}

Endoscopy, for both diagnostic and therapeutic interventions, provides direct visualization and treatment capabilities within the gastrointestinal (GI) tract~\cite{wang2023rethinking,gralnek2022endoscopic,bai2023surgical}. However, traditional endoscopic methods face limitations in precise control, particularly in complex three-dimensional digestive environments. The introduction of capsule endoscopy has alleviated some of these challenges by offering a minimally invasive alternative. Advancements in capsule endoscopy, such as high-resolution imaging, wireless data-enhanced non-invasive diagnostics, and targeted therapies, minimize patient discomfort and risks~\cite{iddan2000wireless,tan2024endoood}. However, without active control, capsule robots remain largely passive, relying on peristalsis for movement, which limits their diagnostic and therapeutic precision. To address these challenges and enhance operator-machine interactions, we propose CapsDT, a customized Diffusion Transformer model for capsule endoscopy robots, drawing inspiration from RDT~\cite{liu2024rdt}.

CapsDT is developed as part of a comprehensive robotic system designed for autonomous capsule endoscopy. This system consists of a 7-DoF robotic arm that holds and controls an external permanent magnet, a capsule robot equipped with an internal magnetic dipole, and a silicone gastric model that simulates the human stomach environment. The objective of this system is to achieve precise control of the capsule robot for diagnostic and therapeutic tasks, enabling efficient and intuitive operator interactions.

Achieving active locomotion in capsule endoscopy is essential for enhancing imaging stability, targeted lesion detection, and controlled rotational maneuvers~\cite{bai2025v}. In this study, we assume that the capsule robot is navigated within the stomach by an external magnetic field, while in the small intestine, it remains passively propelled by gastrointestinal peristalsis. Consequently, automating navigation and motion control for endoscopic imaging presents significant challenges, requiring a Vision-Language-Action (VLA) system to process multimodal sensory inputs and execute precise movements in a dynamic three-dimensional environment~\cite{ng2025endovla}. To achieve this, accurate localization and autonomous path planning are crucial, forming the foundation for downstream therapeutic interventions, including lesion detection, medication administration, and biopsy procedures.

To systematically enhance the autonomous capabilities of capsule robots, we define four progressive endoscopic tasks: basic navigation, rotational control, navigation in fluid-filled environments, navigation and rotation in fluid-filled environments.

CapsDT is specifically designed for these multi-task endoscopic challenges, processing robot parameters as input and generating optimal robotic control outputs. The model undergoes a pre-training phase with fundamental textual instructions and images, followed by action learning, where it directly generates control signals. By leveraging pre-trained Vision-Language-Action models, CapsDT synthesizes visual and linguistic data to generate precise robotic maneuvers, enabling autonomous control in real-world conditions. The key contributions of our work are as follows:
\begin{itemize}
\item[--] We develop an endoscopy capsule robot system, that includes a capsule robot, a 7-DoF Kuka robot arm, and a stomach simulator combined with four distinct endoscopy tasks and task-specific datasets. Kuka robotic arm controls a magnet, creating a controlled magnetic field that actuates the endoscopy capsule robot, enabling it to move. CapsDT on a self-collected multi-task capsule robotics dataset comprising over 1k trajectories, which is one of the most extensive endocopy capsule robot datasets.
\item[--] We present a novel Vision-Language-Action diffusion transformer model, CapsDT, which integrates robotic sensors, visual data, and natural language instructions. This approach effectively bridges linguistic expressions with perceptual cues, enabling the endoscopy robot to generate optimal actions. Our experimental results demonstrate that CapsDT achieves state-of-the-art performance, surpassing baseline models by improving success rates than baseline by 21.25\% across four endoscopic tasks.
\end{itemize}

\section{RELATED WORK}

Robotics in endoscopy address the limitations of traditional diagnostic and therapeutic procedures~\cite{zhang2022deep}. The evolution of endoscopy robotics can be categorized into two primary systems: robot-driven endoscopes and robot-driven instrumentation~\cite{bovskoski2019endoscopy}. Robot-driven endoscopes enhance precision and maneuverability, applied in surgeries like NOTES~\cite{swanstrom2011platform}, ESD~\cite{kume2023flexible}, mucosectomy, and others, where multi-DOF robotic systems have expanded capabilities, allowing complex procedures with greater accuracy and safety. Multi-degree-of-freedom (DOF) robotic systems have enabled complex procedures~\cite{NOTES2011, Flexi2023}. Morino et al. introduced a flexible endoscopic robot for transanal surgery, improving adaptability in confined spaces~\cite{Init2022}. Luo et al. developed snake-like flexible manipulators for gastrointestinal endoscopy, enhancing intraluminal mobility with biomimetic movements~\cite{ANovel2023}. Gruijthuijsen et al. advanced autonomy by creating a framework for robotic endoscope coordination, enabling robot arm manipulation in dynamic environments~\cite{Robot2022}.

Recent innovations explore magnetic actuation for wireless capsule endoscopy (WCE), enabling minimally invasive diagnostics. Traditional Continuous Rotation Magnetic Actuation (CRMA)~\cite{10623502} uses magnetic torque and drag for propulsion but risks intestinal damage due to friction. To address this, Xu et al. introduced Reciprocal Rotating Magnetic Actuation (RRMA)~\cite{xu2021reciprocally}, reducing resistance and preventing tissue trauma through alternating rotation. Our system adopts these RRMA principles for safer navigation.

Computational technologies are also transforming endoscopic robotics. Simsek et al. proposed integrating AI and 3D imaging to improve precision, shorten learning curves, and enable real-time decision-making during procedures~\cite{Futur2023}.  Additionally, advancements in WCE, such as high-resolution imaging, longer battery life, and reliable wireless data, enhance non-invasive diagnostics and targeted therapies, minimizing patient discomfort and risks.

\section{Endoscopy Capsule Robot System}

\begin{figure}[h]
    \centering
    \includegraphics[width=1\linewidth]{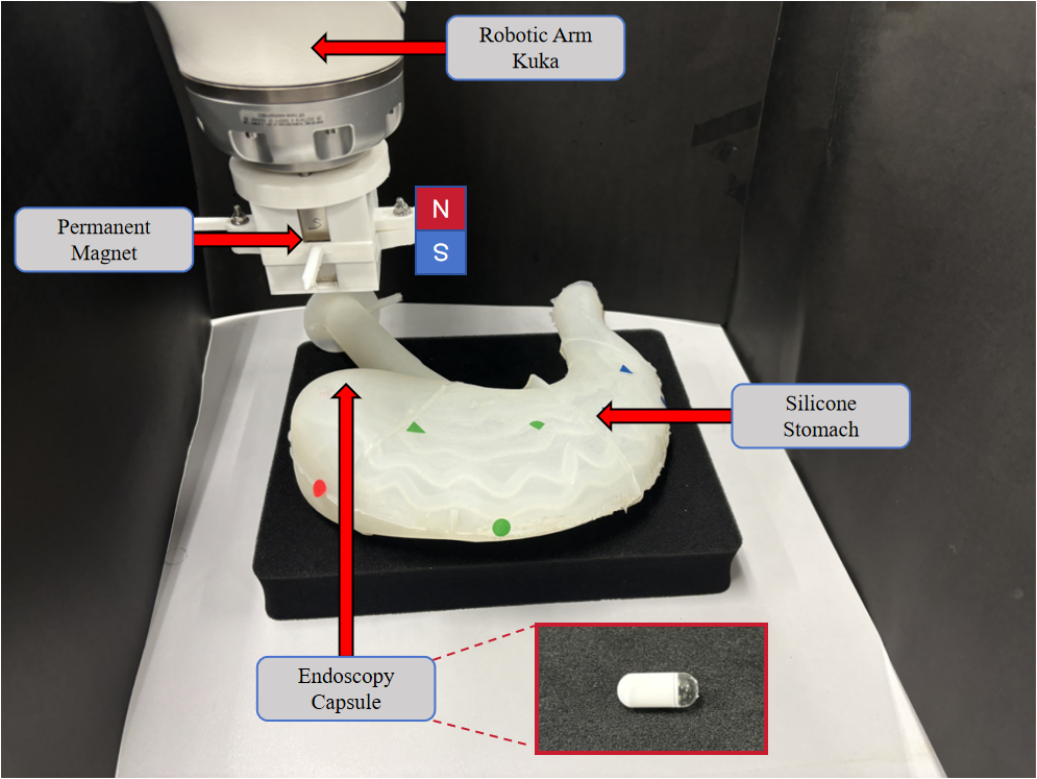}
    \caption{\textbf{Endoscopy Capsule Robot System.} CapsDT generates control commands for the robotic arm to adjust the magnet’s position and orientation, creating a controlled magnetic field that actuates the endoscopy capsule robot, enabling its navigation within the silicone stomach.}
    \label{fig_task}
\end{figure}

\begin{table*}[t]
    \centering
    \caption{\textbf{Endoscopy capsule robot tasks.}}
      \resizebox{\textwidth}{!}{	
    \begin{tabular}{lcc}
        \hline
        TASK NAME&DIFFICULTY LEVELS&EXPLANATION\\
        \hline
        Navigation&low-level&Go through the stomach\\
        Rotation&low-level&Rotate 90° clockwise then 90° counterclockwise\\
        View adjustments & medium-level & Change the capsule camera views in different water lines while navigating \\
        View adjustments with rotation & high-level & Change the capsule camera views in different water lines and rotate 90° clockwise and counterclockwise\\
        \hline
    \label{tab_task}
    \end{tabular}}
    \vspace{-7mm}
\end{table*}

As shown in Fig.~\ref{fig_task}, the system follows a chain-control architecture, where CapsDT generates control commands for the robotic arm. The robotic arm adjusts the magnet’s position and orientation, creating a controlled magnetic field that actuates the endoscopy capsule, enabling its navigation within the silicone stomach.

The proposed model processes sequential visual-linguistic inputs through a unified architecture, where the model ingests several consecutive image frames paired with natural language instructions to predict 7-dimensional action vectors governing the end-effector's motion. These action parameters encode both translational velocities ($\Delta x, \Delta y, \Delta z$ in mm/s) and rotational velocities ($\Delta\theta_{roll}, \Delta\theta_{pitch}, \Delta\theta_{yaw}$ in rad/s), along with a normalized gripper actuation signal $g \in [0,1]$. 

In our system, the permanent magnet is fixed to the end effector of the redundant 7-DOF (Degrees of Freedom) robotic arm. The robotic arm used in this study is KUKA LBR iiwa 7 R800, which can provide high-precision and stable control.
The customized fixture rigidly mounts the permanent magnet onto the robotic arm’s end-effector, establishing a fixed distance $37.0 mm$ between the geometric center of a cubic magnet and the tool center point (TCP) of the 7-DOF manipulator. 
The permanent magnet is a cube with a side length of 50 mm and a magnetic dipole moment of $119.6A·m^2$. The downward-facing south pole of the magnet results in a magnetic moment directed downward. By modifying the end-effector's orientation, the direction of the magnetic moment can be adjusted to achieve the desired magnetic field configuration.
Through forward kinematic analysis based on Denavit-Hartenberg (D-H) parameters, the end-effector's pose (position and orientation) and the direction of the magnetic moment can be mathematically determined. 
In this experiment, an Ankon capsule with dimensions of $26 mm$ × $15 mm$ and a mass of $4.56 g$ is utilized. The capsule has a magnetic dipole moment of $0.126 A·m^2$, enabling controlled motion under an external magnetic field. A silicone gastric model featuring internal rugae structures is utilized to mimic the human stomach environment, providing a realistic setting for capsule motion.

To calculate the magnetic field force on an endoscopy capsule \cite{mahoney2016five}:

\begin{equation}
F_m=\frac{3 \mu_0\|M\|\|m\|}{2 \pi\|r\|^4}
\end{equation}\\
where $\mu_0$ denotes the vacuum permeability, which equals to $4\pi \times 10^{-7} \frac N{A^2}$. The magnetic dipole moments of the permanent magnet and the internal magnet of the endoscopy capsule are represented by the vectors ${M}$ and ${m}$, respectively. The vector ${r}$ represents the relative position between these two magnetic dipoles.

\subsection{Endoscopy Task Dataset and Task Design}\label{data}

The experimental data acquisition process employs a joystick-based motion control interface, with a real-time cartesian direct servo mode ensuring smooth and continuous magnetic field motion.  
 
For data acquisition, the silicone gastric model is secured in a predefined position, and the capsule is initially placed at the esophageal entry point to ensure a consistent starting condition. 
The robotic arm then moves the magnet to its designated starting position before entering real-time control mode. Once activated, the system performs an initial check to verify the joystick input's responsiveness.
Joystick inputs are mapped to the control of the magnet’s pose. By manipulating the joystick, the permanent magnet guides the endoscopy capsule to move in the desired direction and navigate through the silicone gastric model. During this process, the joint angles of all seven joints, as well as the pose (position and orientation) of the permanent magnet, and corresponding timestamps are recorded at a frequency of $10 Hz$. 

In capsule robot manipulation, a specific task usually consists of several components: a skill (e.g., verbs like “detect” or “rotate”), an object (e.g., nouns like “stomach”), a scene (i.e., the environment where the task is completed in the water), and a modality that describes how the skill is performed (e.g., adverbials like “with different water lines”). The details of endoscopy tasks are provided in Table~\ref{tab_task}.

Despite extensive pre-training on large-scale datasets, CapsDT may still encounter limitations in zero-shot generalization to target capsule robots due to the embodiment gap. To address this challenge, we propose the collection of a multi-task endoscopic capsule robot dataset specifically tailored for the target robot, enabling effective fine-tuning. Our methodology involves the creation of a comprehensive multi-source dataset through three distinct approaches: (1) Quantitative robustness: We have compiled 1,000 trajectories, establishing our dataset as one of the most extensive resources currently available for capsule robotics research; (2) Task diversity: The dataset encompasses four complex tasks that represent the majority of manipulation task categories, ranging from basic navigation to advanced rotational maneuvers; (3) Textual augmentation: We employ large language models (LLMs) Deepseek~\cite{liu2024deepseek} to systematically rewrite human-annotated instructions, generating 100 distinct but semantically similar instruction sets to enhance linguistic diversity. Consequently, each data point in our dataset comprises a text instruction with detailed LLM-generated explanations, accompanied by dual visual perspectives from both the gripper and exterior viewpoints. This comprehensive approach ensures a robust foundation for model training and evaluation in capsule robotics applications.

\section{METHODOLOGY}

\begin{figure*}[h]
    \centering
    \includegraphics[width=1\linewidth]{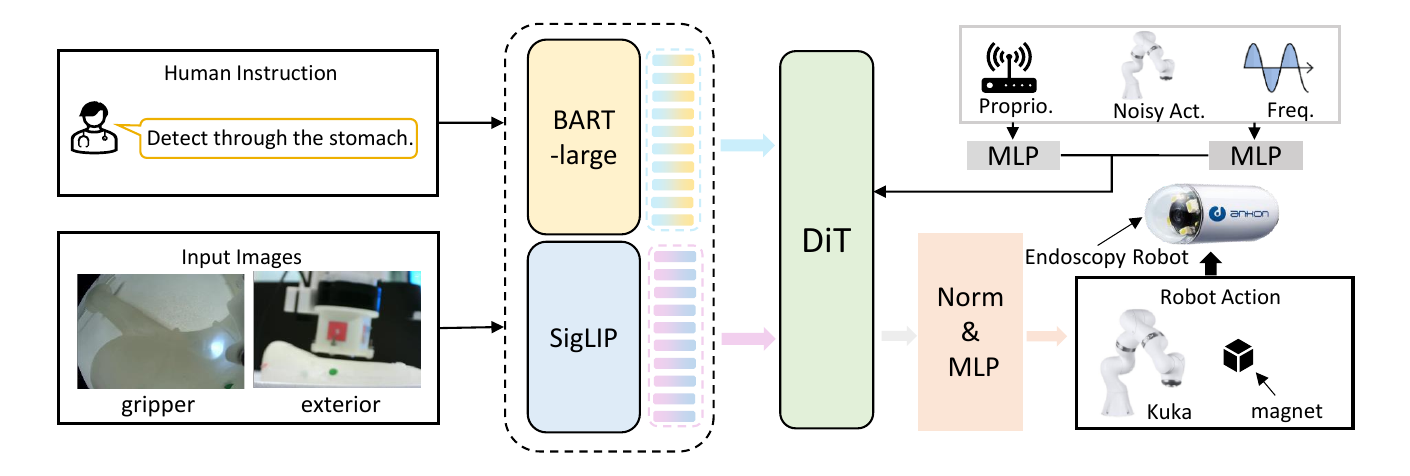}
    \caption{\textbf{CapsDT framework.} For multi-robot training, the action spaces of different robots are integrated into a single action environment. As denoising inputs, the following inputs are used: proprioception, noisy action chunk, control frequency, and diffusion time step; as conditions, picture inputs (a collection of images from the gripper camera and the exterior); and as conditions, language inputs. Denoised action is the output.}
    \label{model}
    \vspace{-6mm}
\end{figure*}
As shown in Fig. \ref{model}, we introduce the Capsule Robot Diffusion Transformer (CapsDT). In this section, we introduce the diffusion model and related architecture. In order to enhance CapsDT's robot manipulation capabilities, we additionally gather an extensive multi-task endoscopic capsule robot dataset for fine-tuning.

\textbf{Diffusion Modeling.} Given the linguistic instruction $\ell$ and observation $o_t$, multi-modality may result in a large number of viable actions $a_t$ to complete the job.  If we model it as a deterministic mapping $(\ell,\boldsymbol{o}_t)\mapsto\boldsymbol{a}_t$ and regress the tuples of $(\ell, o_t, a_t)$ in the training data, the policy will learn the "average" of action modes.  As a result, there may be out-of-distribution actions that are entirely impossible~\cite{chi2023diffusion}, like the arithmetic mean of many modes. The continuous conditional distribution $p(a_t|\ell, o_t)$ is what we decide to model instead. 

However, since picture data varies from the inherent properties of robotic physics variables (i.e., action and proprioception), applying diffusion models to robotic activities poses unique challenges.  Image data often exhibits some degree of temporal and spatial continuity \cite{2019Capturing}, despite being high-dimensional, with changes often taking place incrementally between frames.  Alternatively, they are defined by the nonlinear dynamics of robotic physics quantities \cite{de2012theory} and the potential for high-frequency changes brought about by physical interactions such as collision, constraints, and material properties such as the stomach. Furthermore, the amounts show an unstable numerical range, which is probably caused by extreme readings from faulty sensors.  This highlights the necessity to adapt existing diffusion models to sufficiently account for the instability and nonlinearity of robot data. 

A completely noisy action $a^K_t \sim N (0, I)$ is sampled first when using diffusion policies. It is then denoised using $K \in N+$ denoising steps to produce a clean action sample $a^0_t$ from $p(a_t|\ell, o_t)$:

\begin{equation}\boldsymbol{a}_t^{k-1}=\frac{\sqrt{\bar{\alpha}^{k-1}}\beta^k}{1-\bar{\alpha}^k}\boldsymbol{a}_t^0+\frac{\sqrt{\alpha^k}(1-\bar{\alpha}^{k-1})}{1-\bar{\alpha}^k}\boldsymbol{a}_t^k+\sigma^k\boldsymbol{z},\quad k=K,\ldots,1,\end{equation}

where a noise schedule \cite{nichol2021improved} pre-defines the scalar coefficients $\{\alpha^k\}_{k=1}^K,\{\sigma^k\}_{k=1}^K$.  In this case, $\beta^{k}:=1-\alpha^{k},$ and$\bar{\alpha}^{k-1}:=\prod_{i=1}^{k-1}\alpha^i,\boldsymbol{z}\sim\mathcal{N}(\boldsymbol{0},\boldsymbol{I})\mathrm{~if~}k>1,\mathrm{else~}\bar{\alpha}^{k-1}=1,\boldsymbol{z}=\boldsymbol{0}$.  But before sampling is complete, $a^0_t$ is intractable.  To estimate the clean sample from a noisy one, we choose to employ a learnable denoising network $f_\theta$ with parameters $\theta$:  The left-arrow $f_{\boldsymbol{\theta}}(\ell,\boldsymbol{o_t},\boldsymbol{a_t^k},k)$ is $\boldsymbol{a_t^0}$.  We shall reduce the following mean-squared error (MSE) of denoising in order to train such a network:

\begin{equation}\mathcal{L}(\boldsymbol{\theta}):=\mathrm{MSE}\left(\boldsymbol{a}_t,f_{\boldsymbol{\theta}}(\ell,\boldsymbol{o}_t,\sqrt{\bar{\alpha}^k}\boldsymbol{a}_t+\sqrt{1-\bar{\alpha}^k}\boldsymbol{\epsilon},k)\right),\end{equation}

where $k\sim\mathrm{Uniform}(\{1,\ldots,K\}),\boldsymbol{\epsilon}\sim\mathcal{N}(\boldsymbol{0},\boldsymbol{I}),\mathrm{and~}(\ell,\boldsymbol{o}_t,\boldsymbol{a}_t)$ is sampled from our training dataset. Later in this paper, we will denote noisy action inputs by $\tilde{\boldsymbol{a}}_t:=\sqrt{\bar{\alpha}^k}\boldsymbol{a}_t+\sqrt{1-\bar{\alpha}^k}\boldsymbol{\epsilon}$, in which the superscript of k is dropped for simplicity. Besides, in practice, we prefer to predict a sequence of actions, i.e., an action chunk, in one shot to encourage temporal consistency \cite{chi2023diffusion} and to alleviate error accumulation over time by reducing number of decisions in a task \cite{zhao2023learning}. Specifically, we model $p(\boldsymbol{a}_{t:t+T_a}|\ell,\boldsymbol{o}_t),\mathrm{where~}\boldsymbol{a}_{t:t\boldsymbol{+}T_a}:=(\boldsymbol{a}_t,\ldots,\boldsymbol{a}_{t\boldsymbol{+}T_a\boldsymbol{-}1})$ is an action chunk and Ta denotes the chunk size.

\textbf{Multi-Modal Inputs Encoder.} The structure reflects the multi-modal inputs, meaning that each modality has a notably distinct format and number of dimensions. Multimodal training has faced difficulties as a result. We solve this problem based on RDT~\cite{liu2024rdt} by encoding these many modalities into a single latent space. The encoding techniques are as follows:
1) Physical characteristics of the robot, such as proprioception, action chunk, and control frequency, are represented as low-dimensional vectors. The high-frequency changes may be efficiently captured by MLPs (with Fourier features \cite{tancik2020fourier}), which we utilize to encode them.
2) Both geographical and semantic information may be found in image inputs. We employ SigLIP~\cite{radford2021learning}, an image-text-aligned pre-trained vision encoder, to extract compact representations.
3) We use a Transformer-based language model that has already been trained, BART-large \cite{lewis2019bart}, to encode them. 

Multimodal aspects include structure as well as the varying quantities of information present in various inputs. For instance, photos often generate more tokens after encoding and carry more information than text. Second, the quantity of information included in several inputs of the same modality might vary greatly. For instance, compared to gripper cameras, a robot's external camera provides a more comprehensive picture and more detailed information. In this situation, the model may pick up a shortcut by ignoring the gripper views and concentrating solely on the outside view, which would impair its ability to sense depth. In order to address this problem, we individually and randomly mask each multi-modal input during encoding with a particular probability. This keeps the model from becoming overly dependent on any one input.

\section{EXPERIMENTS}

\begin{figure*}[h]
    \centering
    \includegraphics[width=1\linewidth]{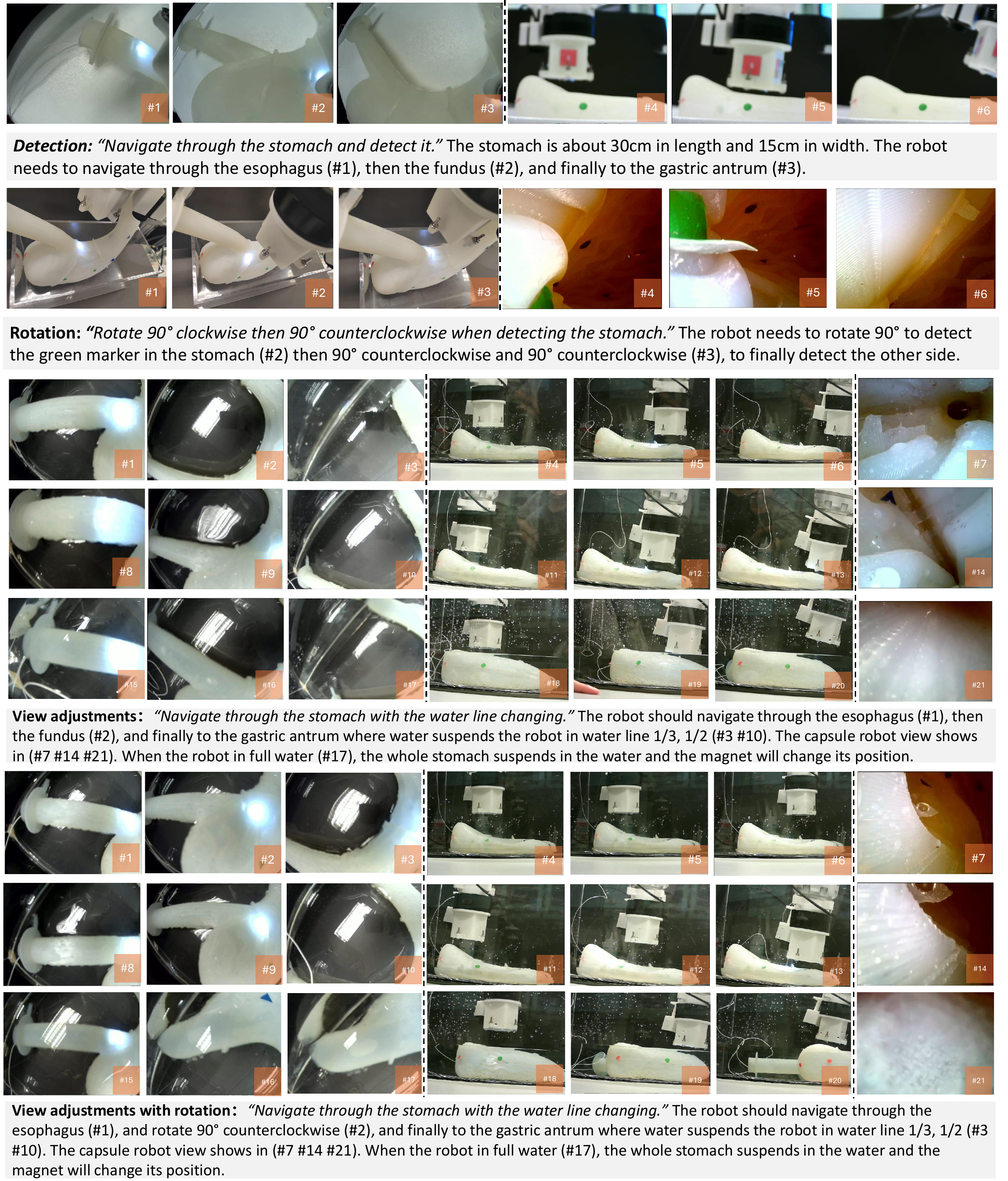}
    \caption{\textbf{Task definitions and visualizations.} For 4 challenging tasks, we describe their language instruction, randomization, and definitions of each sub-task. The figures include the view of the gripper, exterior, and capsule robot. The photos of the gripper and the exterior camera will be the inputs of CapsDT. The lights in the picture realistically represent the position of the capsule robot.}
    \label{2}
    \vspace{-6mm}
\end{figure*}

\subsection{Experiment setting}
\textbf{Tasks.}
To assess CapsDT's generalisability and capabilities, we select four difficult challenges from various angles. These tasks include intricate situations that the model may face in the real world, such as dexterous manipulation and a stomach simulator. Table~\ref{tab_task} shows the dimensions of each task, and Fig.~\ref{2} offers comprehensive explanations and visualizations.

\textbf{Data.}
We use the pre-training and fine-tuning datasets in Sec. \ref{data}. We now list the number of demos related to each task in our fine-tuning dataset. Navigation: 12 demos for stomach simulator; Rotation: 20 demos for rotating 90° clockwise; View adjustments: 10 demos for the water level of one-third, 10 demos for half, and 10 demos for full; View adjustments with rotation: 5 demos for the water level of one third, 5 demos for half, and 5 demos for full.

\begin{table*}[t]
\caption{Quantitative Performance Success Rate(\%). We report the performance of Octo, OpenVLA, and CapsDT for 4 tasks. CapsDT outperforms other baseline models on these endoscopy capsule robot tasks.}
\label{sr}
\centering
\begin{tabular}{lccc|ccc|ccc|ccc|c}
\hline
\multicolumn{14}{c}{Navigation} \\ \hline
& \multicolumn{3}{c|}{Navigate to Esophagus}
& \multicolumn{3}{c|}{Navigate to Fundus}
& \multicolumn{3}{c|}{Navigate to Gastric antrum}
& \multicolumn{4}{c}{Total} \\ \hline
Octo      &   & 33 &   &   & 0 &   &   & 0 &   & \multicolumn{4}{c}{0} \\
OpenVLA   &   & 8  &   &   & 0 &   &   & 0 &   & \multicolumn{4}{c}{0} \\
CapsDT (Ours)    &   & 66 &   &   & 50 &   &   & 25 &   & \multicolumn{4}{c}{\textbf{25}} \\ \hline

\multicolumn{14}{c}{Rotation} \\ \hline
& \multicolumn{3}{c|}{Rotate 90° clockwise}
& \multicolumn{3}{c|}{Back into place} 
& \multicolumn{3}{c|}{Rotate 90° counterclockwise} 
& \multicolumn{4}{c}{Total} \\ \hline
Octo      &   & 55 &   &   & 30 &   &   & 50 &   & \multicolumn{4}{c}{20} \\
OpenVLA   &   & 40 &   &   & 30 &   &   & 50 &   & \multicolumn{4}{c}{15} \\
CapsD (Ours)    &   & 70 &   &   & 50 &   &   & 60 &   & \multicolumn{4}{c}{\textbf{40}} \\ \hline

\multicolumn{14}{c}{View adjustments $|$ 1/3 water line$|$ 1/2 water line $|$ Full water line} \\ \hline
& \multicolumn{3}{c|}{Navigate to Esophagus}
& \multicolumn{3}{c|}{Navigate to Fundus} 
& \multicolumn{3}{c|}{Navigate to Gastric antrum} 
& \multicolumn{4}{c}{Total} \\ \hline
Octo      & 10 & 10 & 10 & 0 & 0 & 0 & 0 & 0 & 0 & \multicolumn{4}{c}{0} \\
OpenVLA   & 20 & 10 &  0 & 0 & 0 & 0 & 0 & 0 & 0 & \multicolumn{4}{c}{0} \\
CapsDT (Ours)    & 80 & 70 & 80 & 50 & 50 & 20 & 20 & 20 & 20 & \multicolumn{4}{c}{\textbf{20}} \\ \hline

\multicolumn{14}{c}{View adjustments with rotation $|$ 1/3 water line $|$ 1/2 water line $|$ Full water line} \\ \hline
& \multicolumn{3}{c|}{Navigate to Esophagus}
& \multicolumn{3}{c|}{Rotate 90° clockwise}
& \multicolumn{3}{c|}{Navigate to Fundus}
& \multicolumn{3}{c|}{Navigate to Gastric antrum} 
& \multicolumn{1}{c}{Total} \\ \hline
Octo      & 40 & 40 & 20 & 40 & 40 & 0 & 0 & 0 & 0 & 0 & 0 & 0 & 0 \\
OpenVLA   & 20 & 20 & 20 & 40 & 20 & 0 & 0 & 0 & 0 & 0 & 0 & 0 & 0 \\
CapsDT (Ours)    & 80 & 80 & 60 & 80 & 80 & 20 & 60 & 60 & 0 & 40 & 20 & 0 & \textbf{20} \\ \hline
\end{tabular}
\end{table*}

\textbf{Model Training and Inference.}
We design CapsDT, which is the first diffusion-based endoscopy capsule robotic model available at the moment. For a week, the model is pre-trained on four RTX 4090 24GB GPUs, resulting in 1000 training iterations. This model is fine-tuned over the course of 24 hours using the same GPUs for 100 steps. We use DPM-Solver++ \cite{lu2022dpm}, a new sampling accelerator of diffusion models, for real-time inference.

\textbf{Baselines.}
We take into account the most sophisticated baselines in robotic foundation models and capsule robotics manipulation, such as Octo \cite{team2024octo} and Action Chunking OpenVLA \cite{kim2024openvla}, in order to thoroughly assess CapsDT. The largest open-source foundation model (7B) that uses discretization modeling is called OpenVLA. The biggest version of Octo, a diffusion-based foundation model, contains just 93 million parameters. 

\textbf{Metric and Hardware.} We employ the success rate as our main metric, which is calculated by dividing successful trials by total trials. Navigation is tested with 5 trials. Rotation is tested with 10 trials. Pour View adjustments are tested with 5 trials for each water line. View adjustments with rotation are tested with 5 trials for each water line. All the tests are performed on the Kuka robot. 

\subsection{Quantitative Results}

Zero-shot generalization to scenes and modalities is possible with CapsDT. CapsDT can still accomplish a high success rate in "View adjustments" and "Rotation" when dealing with shifting water conditions, and its performance is mostly unchanged from that of the observed ones. The other baselines, on the other hand, are unable to do the full work. Even though it has never encountered terms like "one-third" or "full," we can observe from "View adjustments with rotation - 1/3 and full" that CapsDT closely follows instructions by manipulating and changing its position while accurately suspended in the water. CapsDT's great zero-shot generalization is a direct result of extensive pre-training, which exposed it to a wide range of aquatic environment conditions and instructions.

CapsDT can learn new skills using only a few shots. In "view adjustments with rotation", CapsDT has learned new and sophisticated abilities of rotation by few-shot learning, whose action patterns are considerably different from existing skills in different water lines, while the success rate of others is nearly zero. Large-scale pre-training is also responsible for this development. CapsDT can swiftly adjust to different work situations with the use of few-shot learning, which is crucial for real-world applications.

CapsDT can handle sensitive tasks. In contrast to other capsule robots that have caused the robot to veer off course, CapsDT carefully controls the angle while pushing the joystick in the "View adjustments with rotation" task. This is so that the demands of dexterous activities may be met by the action's precision. Our resilient network architecture allows diffusion to explain the spread of nonlinear and multi-modal activities. Furthermore, we note that when the robot is in the full water situation, the stomach floats on the water, so the movement of the capsule robot will influence the stomach. This special situation probably increases Octo's chances of failing. However, thorough pre-training has enhanced CapsDT's ability to identify by assisting its movement in creating a more accurate vision-language representation of the joystick concept.

\subsection{Ablation Study}
To determine the significance of the model size, pre-training, and modeling technique, we have carried out ablation tests. We take into account the variations of: Our CapsDT: the first CapsDT. LLM takes the position of diffusion modeling in CapsDT (Regress) (Llama2-13B~\cite{touvron2023llama}). The deterministic mapping $(\ell,\boldsymbol{o}_t)\mapsto\boldsymbol{a}_t$ is modeled by Llama2. Pretrained CapsDT: CapsDT that hasn't been fine-tuned. Scratch CapsDT: CapsDT without any prior training. During fine-tuning, CapsDT is trained from the beginning. We assess these variations based on three generalizability dimensions in Table \ref{ablation study}. A comparison of baselines and various CapsDT variations is shown in Table \ref{ablation study}.

\begin{table}[t]
    \centering
    \caption{\textbf{Ablation Study.} The success rates (\%) of the original CapsDT and its three variants in different levels of tasks.}
    \begin{tabular}{lccc}
        \hline
        Model&Low-Level&Medium-Level&High-Level\\
        \hline
        Pretrained&11.6&10.0&0\\
        Scratch&9.1&0&0\\
        Regress&6.6&0&0\\
        \hline
        Ours&32.5&20.0&20.0\\
        \hline
    \label{ablation study}
    \end{tabular} 
    \vspace{-7mm}
\end{table}

Table \ref{ablation study} shows a considerable drop in performance without any of these elements, highlighting the necessity of our contributions.  In particular, CapsDT (scratch) performs poorly on objects and the simulator, indicating that pre-training data is necessary for generalization.

\section{CONCLUSIONS}

By creating the Robotics Diffusion Transformer (CapsDT), a diffusion-based foundation model for language-conditioned visuomotor imitation learning, we addressed the issues of data scarcity and increased manipulation complexity in generalizable capsule robotics manipulation. Our model was refined using a self-collected capsule robot dataset and pre-trained on this self-made dataset. We also present an endoscopic capsule system that consists of a stomach simulator, a 7-DoF Kuka robot arm, and a capsule robot. This system can achieve precise control of the
capsule robot for endoscopy inspection tasks, enabling efficient and intuitive operator interactions, achieving state-of-the-art performance in real-world simulation manipulation.

\bibliographystyle{IEEEtran}
\bibliography{reference}

\addtolength{\textheight}{-12cm}   





\end{document}